\pdfoutput=1

\documentclass[11pt]{article}

\usepackage{acl}
\usepackage[utf8]{inputenc}
\usepackage{csquotes} 
\usepackage{amsmath, amssymb} 
\usepackage{float}
\usepackage{algorithm}
\usepackage[noend]{algpseudocode}
\usepackage{textcomp,latexsym} 
\usepackage{color} 
\usepackage{graphicx}
\usepackage{paralist}

\usepackage{times}
\usepackage{latexsym}

\usepackage[T1]{fontenc}

\usepackage{microtype}

\title{Lex2Sent: A bagging approach to unsupervised sentiment analysis}

\author{Kai-Robin Lange \and Jonas Rieger \and Carsten Jentsch\\
Department of Statistics, TU Dortmund University, 44221 Dortmund, Germany\\
\texttt{\{kalange, rieger, jentsch\} @statistik.tu-dortmund.de}\\}

\begin{document}
\maketitle
\begin{abstract}
Unsupervised text classification, with its most common form being sentiment analysis, used to be performed by counting words in a text that were stored in a lexicon, which assigns each word to one class or as a neutral word. In recent years, these lexicon-based methods fell out of favor and were replaced by computationally demanding fine-tuning techniques for encoder-only models such as BERT and zero-shot classification using decoder-only models such as GPT-4. In this paper, we propose an alternative approach: Lex2Sent, which provides improvement over classic lexicon methods but does not require any GPU or external hardware. To classify texts, we train embedding models to determine the distances between document embeddings and the embeddings of the parts of a suitable lexicon. We employ resampling, which results in a bagging effect, boosting the performance of the classification. We show that our model outperforms lexica and provides a basis for a high performing few-shot fine-tuning approach in the task of binary sentiment analysis.
\end{abstract}

\section{Introduction}
Most commonly, text classification is performed in a supervised manner by using a previously labeled data set to train a learning-based model to predict the sentiment of unlabeled documents. When a labeled data set is not available, an unsupervised labeling approach is useful to provide valuable initial information for an active learning approach or to label the texts right away, when a near-perfect classification is not strictly necessary. However, such unsupervised models often require financial backing or a high performing GPU to use on a large data set.

In this paper, we propose Lex2Sent, a model mainly designed for sentiment analysis, that can however be used for any binary text classification problem, where external resources in the form of lexica are available. We will thus define the model for any arbitrary binary classification. Lex2Sent uses text embedding models to estimate the similarity between a document and both halves of a given binary lexicon. These distances are calculated for multiple resampled corpora and are aggregated to achieve a bagging-effect. As Doc2Vec models are usually trained on the CPU, the method demonstrated here can be fully realized in low hardware resource environments that do not have access to a GPU or the financial means to let commercial models such as GPT label thousands of documents. As the Lex2Sent's architecture is not dependent on the language of choice, it can also be used in other languages than English, including low resource languages for which no powerful language models are available. To demonstrate that the results are generalizable, we compare them to the ones of traditional lexicon methods on three data sets with distinct characteristics. To assess the performance to the modern unsupervised classification state of the art, we compare Lex2Sent's results to GPT-3.5 on one data set. We also extend this active learning approach by fine-tuning a RoBERTa model on a sufficient subset of the labels predicted by Lex2Sent. This can be seen as an initial starting point for active learning approach.

The paper is structured as follows. In \autoref{Related}, we discuss previous approaches to text classification and research on resampling techniques for texts. \autoref{grid} introduces our classification model by describing the Doc2Vec model, the unsupervised labeling approach and the resampling procedure used. The data sets and lexica used are specified in \autoref{Data}. In \autoref{eval}, the classification rates of Lex2Sent are compared to lexicon methods and the performance of Chat-GPT. We also show that we can use the results of Lex2Sent for an initial fine-tuning of a pre-trained language model in few-shot setting. In \autoref{Summary}, we conclude and give an outlook to further research.
\section{Related Work}
\label{Related}
When little to no labeled data is available, usually text classification is performed in one out of three ways. That is, by using either traditional lexicon methods, decoder-only models like GPT or parameter efficient fine-tuning methods to fine-tune pre-trained language models. 

Traditionally, researchers used lexica/dictionaries that were meant to substitute the missing supervised label information by external information. For sentiment classification, such lexica contain both a list of positive and negative words, which could simply be counted within a text. Commonly used lexica include VADER \cite{VADER}, Afinn \citep{Afinn}, Loughran-McDonald \cite{Loughran}, the WKWSCI lexicon \cite{WKW} and the Opinion lexicon \cite{Opinion}. Even within a specific task such as binary sentiment classification, these lexica are often designed for a specific use case. For instance, the Loughran-McDonald lexicon is designed for economic text data, while VADER is designed for social media data. \citet{SpeakGer} perform a sentiment analysis of German political speeches and use Lex2Sent with a lexicon base specifically designed for German political text data \cite{rauh}.\\
This method is very resource-savvy, but yield worse performance than the other two methods. Nowadays, lexica are usually only used in low hardware resource environments or by researchers of social science disciplines, because they are white-box algorithms that are easy to interpret.

Alternatively, GPT-4 \cite{GPT} or any other large language model (e.g. Llama 2 \cite{Llama}, Mixtral \cite{Mixtral} or Jamba \cite{Jamba}) can classify any document in a zero-shot manner due to their language understanding capabilities. Using GPT-4 or GPT-3.5 for large corpora requires financial backing not everyone has access to though and similarly, open source large language models need a GPU with large vram.

Lastly, a pre-trained Transformer model like BERT \cite{BERT} or RoBERTa \cite{RoBERTa}, that was additionally fine-tuned on the task at hand, might help when a GPU is available, that cannot handle a large language model. This however yields the downside of using classification rules that are not based on the texts the model is meant to be used on. Instead, the model might carry a bias from a different subject over to the classification: the sentiment of a text might be based on completely different clues based on whether the text is a political speech or a social media post. This can be avoided by fine-tuning the model oneself, which, in turn, needs labeled data. To reduce the amount of data needed, active learning \cite{AL} is increasingly being used in combination with few-shot learning techniques. Parameter-efficient fine-tuning \cite[PEFT,][]{PEFT} uses adapter methods such as Low Rank Adaption \cite[LoRA,][]{LoRA} to fine-tune language models with fewer training parameters than usual, and is thus suited to fine-tune on few-shot examples to achieve adequate results. Pattern exploiting training \cite[PET,][]{PET} uses the language understanding capabilities of language models to its advantage by ``explaining'' the task to the model. As \citet{Petapter} show, such methods can even be effectively combined into one.

Contrary to these approaches, we propose a fully unsupervised approach that can be used in low hardware resource environments, in which no access to a GPU is available and where there is no financial backing to let commercial models like GPT label thousands of documents. We do this by employing CPU-based embedding algorithms that leverage external information using lexica and are further improved by resampling, resulting in a bagging effect. Improving embedding-based text classification with the help of lexica has been explored by \citet{Shin}, \citet{POLAR1} and \citet{POLAR2}, but neither analyze a combination of embeddings and lexica for unsupervised analysis.

\citet{augmentation} use resampling to improve the performance of supervised sentiment models by resampling words with certain probabilities based on their tf-idf-score or by translating the original document into another language and then translating it back to the original language. Similar augmentations can be performed with the nlpaug-package \cite{nlpaug}. This allows the user to, for instance, use embedding models, be it static models like Word2Vec \cite{Skip} or contextual masked language models like BERT \cite{BERT}. These types of data augmentation and resampling are most often used as additional training data for the embedding models and supervised methods. In this paper, instead of resizing the training set, we create multiple different training sets, on which one embedding model is trained each. Aggregating the information from these models into one combined classifier creates a bagging effect, improving the classification rate. Furthermore, we investigate the advantage of using such augmentation and resampling techniques in an unsupervised setting.

The procedures used by \citet{augmentation} and \citet{nlpaug} do however change the existing vocabulary. They either change the vocabulary by back-translation or resampling the document dependently from other documents due to the tf-idf-scoring or even introduce completely new words that are not part of the corpus at all by changing words based on similiar words in a given embedding space. This might be counter-productive for an unsupervised analysis, as the texts are not used as a training data set, but are supposed to be evaluated themselves. Changing the vocabulary might introduce a bias and hinder the classification performance, as the external information provided to classify the texts is given by the lexica, which are essentially word lists and thus more likely to work accurately to work with unchanged vocabulary. The resampling procedures in this paper are instead based on those employed by \citet{Rieger}, who used resampling procedures to analyze the statistical uncertainty of the topic modeling method Latent Dirichlet Allocation. We chose those procedures, as they augment or resample the texts independently from another and do not add new words to the vocabulary.
\section{Lex2Sent}
\label{grid}
In this section we propose Lex2Sent, a bagging model for unsupervised sentiment analysis. Lex2Sent is published as a Python package. The code can be found on GitHub\footnote{\url{https://github.com/K-RLange/Lex2Sent}}.
\subsection{Lexica}
To perform unsupervised text classification, lexica can be used to interpret the words in a text without the need for previously labeled documents of a similar corpus, as they provide external information. This information is provided in the form of key words, which a lexicon assigns to a certain class.

For our analysis, we use binary lexica, that are used to separate words between two disjoint classes $A$ and $B$. Such a lexicon assigns a value from an interval $[-s,s]$ with $s\in\mathbb R^+$ to all words, while assigning the value $0$ to all neutral words. It assigns positive values to all words it deems to belong to class $A$ and negative values to all words, it deems to belong to class $B$. To enable some words to have a larger weight during the classification process, lexica might give words different values. For instance, the word ``fantastic'' might receive a higher score than the word ``good'' when using a sentiment analysis lexicon, as it conveys an even stronger positive emotion. We modify such a binary lexicon to consist of two halves, one for each of the two classes. These halves are defined as lists of words in a way that each word that belongs to either class $A$ or class $B$ occurs exactly once in its respective half. Only neutral words are not assigned to a half. This enables the use of lexicon-based text embeddings.

As we use static embeddings, a key word's embedding is not changed, even if it is negated in a document. To incorporate the concept of negations into Lex2Sent, we merge negations with the following word during preprocessing. The term \enquote{not bad} is thus changed to \enquote{negbad}. \enquote{negbad} is then added to the opposite lexicon half of the word ``bad'', so that Lex2Sent can interpret it correctly.
\subsection{Lexicon-based text embeddings}
Instead of looking only at key words, text embeddings can be used to analyze semantic similarities to other words. This enables us to identify the class of a text using words that are not part of the lexicon.

Text embedding methods create an embedding for each document, which represents the document as a real vector of some fixed dimension $q$. They are created using the word embeddings of all words in the current document and can be interpreted as an \enquote{average} word embedding. We thus interpret the text embedding of a lexicon half as an average embedding of a word of its respective class. Calculating the distance between the embedding of a document in the corpus and the embedding of a lexicon half is used as a measure of how similar a given document is to a theoretical document that is the perfect representation of that class.

As an alternative to the approach mentioned above, we also looked at the average distance of a document's text embedding to all word embeddings of the sentiment words that appear in the document itself, to analyze only its difference to the parts of the lexicon that are part of the document. However, this yielded a classification rate that is comparable to the traditional lexicon classification itself and does not provide substantial improvement over it. It will thus not be further reported in this paper.

The distance is calculated using the cosine distance
\[\text{cosDist}(a,b)=1-\frac{\sum^q_{i=1}a_ib_i}{\sqrt{\sum^q_{i=1}a_i^2}\cdot\sqrt{\sum^q_{i=1}b_i^2}}\]
for two vectors $a=(a_1,\dotsc, a_q)^T\in\mathbb R^q$ and $b=(b_1,\dotsc, b_q)^T\in\mathbb R^q$ \cite{cos}.

For our purposes of classifying documents into two classes, let $\textit{A}_d$ be the cosine distance of a text embedding of a document to the text embedding of the positive half of a sentiment lexicon and $\textit{B}_d$ be the cosine distance to the negative half. Then, the larger (smaller) the value
\[\textit{diff}_d=\textit{B}_d-\textit{A}_d\]
is for a document $d$, the more confident the lexicon-based text embedding method is, that this document $d$ in fact belongs to class $A$ ($B$). 

This method can be performed using any text embedding model in combination with any lexicon that enables a binary classification task. In this analysis, we choose Doc2Vec \cite{D2V} as the baseline text embedding model and analyze texts for their sentiment.
\subsection{Doc2Vec}
\label{D2V}
Doc2Vec \cite{Doc2Vec} is based on the word embedding model Word2Vec \cite{Skip}, which assigns similar vectors to semantically similar words by minimizing the distance of a word to the words in its context.

Since word embeddings are not sufficient to classify entire documents, the model is extended to text embeddings. A Doc2Vec model, using the Distributed Memory Model approach, uses a CBOW architecture \cite{Skip} in which a document itself is considered a context element of each word in the document. The distance of the document vector to each word vector is minimized in each iteration, resulting in a vector that can be interpreted as a mean of each of its words. According to \citet{Doc2Vec}, these text embeddings outperform the arithmetic mean of word embeddings for classification tasks. In this paper, we use the Doc2Vec implementation of the gensim package in Python \cite{gensim}. 

Formally, we consider $D$ documents and denote by $N_d$ the number of words in document $d\in\lbrace 1,\dotsc,D\rbrace$. Further, for $i\in\lbrace 1,\dotsc,N_d\rbrace$, let $w_{i,d}$ be the $i$-th word in document $d$ and $w_{\text{doc}}$ denote the document under consideration. To give larger weight to words that follow up on another than words that are far away from another, the window size is varied during training. For a Doc2Vec model, we denote by $K$ the maximum size of the context window. For every word the effective size is then sampled from $\lbrace 1, \dotsc,K\rbrace$ and is denoted as $k_{n,d}$. With these windows, the log-likelihood
\[\sum^{N_d-K}_{n=K}\text{ln}\left(p(w_{n,d}\vert w_{n-k_{n,d},d},\dotsc,w_{n+k_{n,d},d},w_{\text{doc}})\right)\]
is maximized for the documents $d=1,\dotsc,D$ using stochastic gradient descent. $p(\cdot\vert\cdot)$ is calculated by the resulting probabilities from a hierarchical softmax \cite{Skip}. 

We also investigated, if the Lex2Sent method would work when using a pre-trained language model, in this case RoBERTa-large \cite{RoBERTa}, as the embedding-backend. For this, we used the CLS-vectors of the lexicon halves and the documents to create lexicon-based text embeddings (similar to \citet{POLAR2}). These results underperformed compared to Doc2Vec though, as they showed a bias for one of the two classes.
\subsection{Text resampling}
\label{resamp}
Word and text embedding models analyze the original text structure to create similar word embeddings for semantically similar words. We assume that lexicon-based text embeddings need an \enquote{optimal text structure} to identify the class of the text in the most efficient manner. Suppose a text contains a key word that is a strong indicatior for the classification task at hand and contained within the lexicon used. The location of such key words can be biased by the type of text. For instance, when analyzing reviews for their sentiment, most key words are located in the last third of the text, as this part draws the conclusion to the review. By resampling the text, we relocate the key words evenly within texts. In theory, this enables vocabulary that occurs more often in texts of a specific sentiment that is not part of any sentiment lexicon, such as topic-specific vocabulary, to be used for labeling texts more efficiently while training Doc2Vec.

We leverage resampling procedures proposed by \citet{Rieger}, who used them to analyze the uncertainty of the Latent Dirichlet Allocation. Instead of analyzing our methods uncertainty, we use these procedures to create optimal text structures and create a bagging effect. For this, we interpret the original text as a bag of words in which words are drawn independently with replacement like observations when creating a bootstrap sample \citep{Efron} or independently without replacement, resulting in a permutated text. We call these procedures BW (Bootstrap for Words) and BWP (Bootstrap for Word Permutation), respectively. We analyzed additional procedures, such as resampling sentences as a whole or resampling words only within sentences and variations of those, but these generally yielded lower classification rates than the procedure described above.
\subsection{Bagging}
\label{Avg}
In this subsection, we describe a technique to aggregate multiple text embeddings for the purpose of unsupervised sentiment analysis. In combination with resampled texts, this can be seen as a bagging method for unsupervised text classification \cite{bagging}. Every text structure has an effect on the classification of lexicon-based text embeddings, as differing syntax and vocabulary change the resulting embeddings. However, identifying whether the texts already have an \enquote{optimal} structure is a difficult task, as this is an abstract concept that is not trivial to formalize. Instead of relying on the original texts' structure, resampling enables the possibility to create an arbitrary number of artificial texts. If we aggregate these text embedding models, they do not have to label a document correctly for one text structure (that is the original text), but instead only have to label a document correctly on average based on multiple differently structured texts. This aggregation also balances out the randomness of generating samples and the negative effect of missing out on a crucial word within documents in one resampling sample, as it will probably appear in other samples.

The aggregation is performed by calculating an average \textit{diff}-vector using $B$ resampling iterations. Let $\textit{diff}{\,}^b_d$ be the $d$-th element of the \textit{diff}-vector for the $b$-th lexicon-based text embedding model with $d=1,\dotsc,D$. Then \[\textit{diff}{\,}^\text{mean}_d :=\frac{1}{B}\sum^B_{b=1}\textit{diff}{\,}_d^b\]
defines the $d$-th element of the averaged \textit{diff}-vector.
\subsection{Algorithm and Implementation}
In training, the algorithm iterates over a grid, calculating models for different training epochs, context window sizes and embedding dimensions. For our application, we use a 3$\times$3$\times$4-grid, which turns out to be sufficiently beneficial in application while remaining computationally feasible. The parameters are chosen from an equidistant set over reasonable parameter choices (see \autoref{shuf_pseudo} for the parameter choices). The grid can be adjusted according to the practitioner's problem at hand. For instance, a smaller grid is faster to train, but a larger grid will lead to more robust results. In each iteration, the parameter combination for the Doc2Vec model is chosen from the grid and the corpus is resampled. The resampled documents are sorted ascendingly by their respective absolute lexicon score. Then we train a Doc2Vec model and calculate the \textit{diff} vector for all iterations. The classification task is performed by using the component-wise arithmetic mean of all the $36$ \textit{diff}-vectors. The algorithm is described as pseudocode in \autoref{shuf_pseudo}. 

Given a classifier $x = (x_1,\dotsc,x_D)\in\mathbb R^D$, the document with the index $d\in \lbrace 1,\dotsc, D\rbrace$ is labeled
\[\text{label}_d = \begin{cases} \text{class A}, &x_d - t<0\\ \text{class B}, &x_d - t > 0\\ \text{at random}, &x_d - t = 0\end{cases},\quad t\in\mathbb R\]
for some threshold $t\in\mathbb R$. This might be $t=0$ or the empirical quantile $t = x_{(p)}$, where $p$ is the estimated proportion of texts of class $B$ based on a-priori knowledge. In the analyses of this paper, we assume to have no a-priori knowledge of the distribution of class labels, so we use $t=0$.
\begin{algorithm*}
\caption{Lex2Sent}\label{shuf_pseudo}
\begin{algorithmic}[1]
\Procedure{Lex2Sent(texts, threshold, lexicon, resampling)}{}
\State classifier $\gets$ [0] $*$ length(texts)
\For{(epoch, window, dim) in Grid = ($\lbrace 5,10,15\rbrace, \lbrace 5, 10, 15\rbrace, \lbrace 50, 100, 150, 200\rbrace$)}
\State resampled\_texts $\gets$ resampling(texts)
\State sorted\_resampled\_texts $\gets$ sort(resampled\_texts, lexicon)
\State model $\gets$ Doc2Vec(sorted\_resampled\_texts, epoch, window, dim)
\State emb $\gets$ lexicon\_based\_text\_embeddings(model, resampled\_texts)
\For{i in 1:length(emb)}
\State classifier[i] $+=$ emb[i]
\EndFor
\EndFor
\State \textbf{return} label\_by\_threshold(non\_resampled\_texts, classifier, threshold)
\EndProcedure
\end{algorithmic}
\end{algorithm*}

\section{Data sets and lexica}
\label{Data}
In this section, the two sentiment lexica and three data sets used to evaluate Lex2Sent are described.
\subsection{Data sets}
The three data sets considered in this paper are chosen to cover texts with distinct features. The iMDb data set consists of a large corpus with long documents and a strong sentiment compared to the other two data sets. The Airline dataset is more than four times smaller and the documents themselves are also shorter. The Amazon data set represents an intermediate case between these two data sets.

The texts are tokenized and stop words as well as punctuation marks and numbers are removed. Lemmatization is performed to generalize words with the same word stem, if the original word from the text does not already appear in the lexicon. The mentioned methods and stop word list are part of the Python package \textit{nltk} \cite{nltk}.
\paragraph{iMDb data set}
The iMDb data set consists of $50,000$ user reviews of movies from the website \url{iMDb.com}, provided by Stanford University \cite{imdb}. These are split into $25,000$ training and test documents, each containing $12,500$ positive and negative reviews. After preprocessing, each document in the data set is $120.17$ words long on average.
\paragraph{Amazon Review data set}
The Amazon data set is formed from the part of the Amazon Review Data which deals with industrial and scientific products \cite{Amazon}. All reviews contain a rating between one and five stars. Reviews with four or five stars are classified as positive and reviews with one or two stars are classified as negative. We removed reviews with a rating of three stars from the data set because the underlying sentiment is neither predominantly negative nor positive. In addition, we filtered out reviews consisting of less than $500$ characters. Out of the remaining documents, $52,000$ documents are split into $26,000$ training and $26,000$ test documents, which are formed from $13,000$ positive and $13,000$ negative documents each. The average length of all documents in the training corpus is $85.51$ words after preprocessing.
\paragraph{Airline data set} 
The third data set consists of $11,541$ tweets regarding US airlines and was downloaded from Kaggle \cite{Kaggle}. The tweets are categorized into positive or negative tweets -- 3099 neutral tweets are deleted to be able to use the data set for a two-label-case. We split this data set in half into a training and test set. The training set ultimately contains 5570 documents. On average, each document of the training set contains $10.60$ words after preprocessing. In comparison to the other two data sets, where the labels are evenly split, in the Airline data set only $1386$ and thus $24.02\%$ of the documents are labeled positive.
\subsection{Lexica}
To demonstrate that the performance is not dependent on the lexicon chosen as a base, we show the performance for three lexica: The Opinion Lexicon \cite{Opinion} is used to represent as a review-specific sentiment lexicon, while the WKWSCI lexicon \cite{WKW} is chosen as multiple-purpose lexicon. The Loughran-McDonald \cite{Loughran} lexicon was designed for economic texts and not for reviews, hence it represents the case in which a lexicon is used in a sub optimal domain. To make sure that Lex2Sent not only outperforms these two lexica, we also observed the classification rate when using VADER \cite{VADER} or Afinn \citep{Afinn} lexicon in the traditional way and compare these results to the one of Lex2Sent in \autoref{lexBase}.

We added four amplifiers and ten negations to improve the classification. If an amplifier occurs before a key word, its value is doubled and if a negation occurs, it is multiplied by $-0.5$. For traditional lexicon methods, the classifier is created by summing up the values of all words within a text.
\begin{table*}[t]
\centering
\caption{Average classification rate in percent of a WKWSCI-based Lex2Sent in comparison to the best lexicon method (in brackets), split into whether the fixed or proportion threshold is used}
\label{fin_tab}
\begin{tabular}{|c||c|c|c|c|}
\hline
& \multicolumn{2}{c|}{WKWSCI-based Lex2Sent} & \multicolumn{2}{c|}{Lexicon with the highest classification rate}\\
\hline 
threshold & by proportion & 0 & by proportion & 0 \\ 
\hline \hline
iMDb & 80.93 & 80.01 & 76.82 (TextBlob) & 73.32 (Opinion Lexicon) \\ 
\hline 
Amazon & 77.08 & 76.83 & 71.91 (VADER) & 69.28 (Opinion Lexicon) \\ 
\hline 
Airline & 79.11 & 72.42 & 82.05 (VADER) & 68.33 (Opinion Lexicon) \\ 
\hline 
\end{tabular} 
\end{table*}

\section{Evaluation}
\label{eval}
The classification rates of Lex2Sent in this section are determined by evaluating 50 executions to observe the method's randomness and to get a metric for the average performance.

\autoref{fin_tab} displays the average classification rates of a WKWSCI-based Lex2Sent and the classification rate of the best performing sentiment lexicon for each data set, split by the classification-threshold used. The WKWSCI-lexicon is chosen as a basis for Lex2Sent as it is a multiple-purpose lexicon. Lex2Sent outperforms every of the 6 observed lexica on all three data sets when using the threshold 0, as it would usually be done in an fully unsupervised setting without a-priori knowledge. It also outperforms the lexica in two out of three cases in which the exact proportion of positive to negative documents is assumed to be known. Here it is only outperformed by VADER on the Airline data set, which is likely because this data set consists of short documents which do not give the Doc2Vec models much context to train on per document.

While Lex2Sent outperforms these lexica, it does not outperform Chat-GPT. \citet{Laskar} report that GPT-3.5 (\texttt{text-davinci-003}) reaches an 91.9\% classification rate on the iMDb data set. While it is not known, if GPT-3.5 has seen this data set and its labels during training and it thus might have an unfair advantage by knowing the correct results \cite{contam}, due to its generally high performance on unsupervised classification tasks, we can assume that it will outperform Lex2Sent, at least on most data sets. Lex2Sent does yield the advantage of not requiring financial backing to analyze large data sets though. Only a CPU is needed.

\begin{table*}[t]
\caption{Average classification rates in percent of a WKWSCI-based Lex2Sent on subsets of the original data sets for the fixed threshold $0$}
\label{tabsubset}
\centering
\begin{tabular}{|c||c|c|c|c|}
\hline 
subsample size & 100\% & 50\% & 25\% & 10\% \\ 
\hline \hline
iMDb & 80.01 & 79.73 & 79.43 & 78.88 \\ 
\hline 
Amazon & 76.83 & 75.71 & 73.79 & 68.86 \\ 
\hline 
Airline & 72.42 & 72.73 & 69.74 & 46.21 \\ 
\hline 
\end{tabular} 
\end{table*}
\subsection{Different resampling procedures} 
In this section, we investigate the effect of different resampling procedures on the performance of Lex2Sent. We examine the results of a WKWSCI-based Lex2Sent using either one of the resampling procedures defined in \autoref{resamp} or no resampling at all for the iMDb data set. Additionally we investigate the classification rate when using texts sorted by their absolute lexicon value (key words grouped at the end of a text). This serves as an ablation analysis to distinguish the effects of resampled, natural and sub optimal text structures (sorted texts). 
\begin{figure}[H]
\centering
\includegraphics[width=0.5\textwidth]{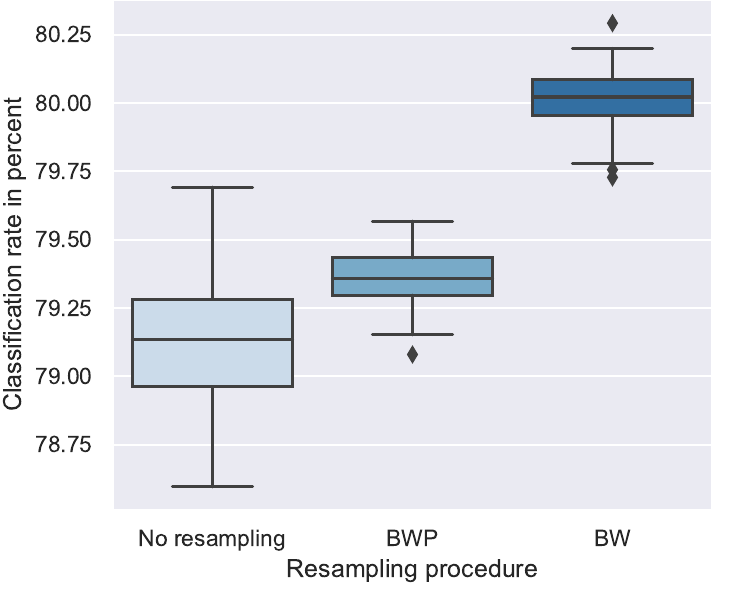} 
\caption{Results of the WKWSCI-based Lex2Sent on the iMDb data set for different resampling procedures}
\label{resComp}
\end{figure}

In comparison to the classification rates displayed as boxplots in \autoref{resComp}, this subotimal text structure results in a strongly decreased classification rate of $71.00\%$, which is in line with our interpretation of \autoref{resamp}. The bagging-effect is visible for both procedures, as using either results in higher classification rates for the iMDb data set, with BW yielding the best performance. The method's stability is also increased, as the classification rates are more consistent, which can be seen by comparing the size of the respective box plots. Similar results (not reported) also occur for the other two data sets. For the rest of this paper, all further results are thus reported for Lex2Sent using BW resampling.
\subsection{Evaluation on smaller corpora}
\begin{table*}[t]
\caption{Average classification rates in percent of Lex2Sent with a WKWSCI-, Loughran McDonald- or Opinion Lexicon-base for the fixed threshold $0$, compared to the rates of the traditional lexicon method on the same lexicon}
\label{TabLexBase}
\centering
\begin{tabular}{|c||c|c|c|c|c|c|}
\hline 
\rule[-1ex]{0pt}{2.5ex}  & \multicolumn{2}{c|}{WKWSCI} & \multicolumn{2}{c|}{Opinion Lexicon} & \multicolumn{2}{c|}{Loughran McDonald} \\ 
\hline 
\rule[-1ex]{0pt}{2.5ex}  & Lex2Sent & lexicon & Lex2Sent & lexicon & Lex2Sent & lexicon \\ 
\hline \hline
\rule[-1ex]{0pt}{2.5ex} iMDb & 80.01 & 70.10 & 78.43 & 73.37 & 70.73 & 61.22 \\ 
\hline 
\rule[-1ex]{0pt}{2.5ex} Amazon & 76.83 & 65.15 & 77.68 & 69.28 & 69.27 & 61.32 \\ 
\hline 
\rule[-1ex]{0pt}{2.5ex} Airline & 72.42 & 63.29 & 71.96 & 68.33 & 72.06 & 53.18 \\ 
\hline 
\end{tabular} 
\end{table*}

As Lex2Sent requires training to accurately represent words with embeddings, it is important to determine how large a corpus needs to be for it to provide sufficient results. To analyze this, we evaluate Lex2Sent for subsamples of each data set. These include 10\%, 25\% or 50\% of the original documents. The results of 50 repetitions are displayed in \autoref{tabsubset}. The classification rates decrease for smaller corpora except for the Airline data set, in which it is slightly higher when examining only 50\% of the data set. On the iMDb data set, Lex2Sent outperforms all lexica, even when using just 10\% of all documents. On the Airline and Amazon data sets, the classification rate of Lex2Sent decreases to a larger extend for smaller subcorpora. This is likely caused by the short documents in these data set and indicates that it is meaningful to use Lex2Sent on smaller data sets if the documents themselves are long enough to train accurate embeddings. 
\subsection{Different lexicon-bases for Lex2Sent}
\label{lexBase}
So far, we focused on the WKWSCI-based Lex2Sent. In this section, we evaluate, how sensitive Lex2Sent is regarding its lexicon-base and if it improves the classification rate of other lexica as well. For this we compare it to Lex2Sent models based on the Opinion lexicon as well as the Loughran-McDonald lexicon. The average classification rates are displayed in \autoref{TabLexBase}. Lex2Sent improves the rates of all three lexica on all data sets. While WKWSCI is a general-purpose lexicon, the Opinion Lexicon is designed to analyze customer reviews. This specialization also affects Lex2Sent, as the Opinion Lexicon-based Lex2Sent outperforms every lexicon on every data set as well as the WKWSCI-based Lex2Sent on the Amazon data set, which consists of product reviews. Similarly, we see that Lex2Sent can improve the performance of a lexicon designed for a different domain, as it increases the classficiation rate for the Loughran-McDonald lexicon by at least 7.95 percentage points on all data sets. We recommend to use a general-purpose lexicon like WKWSCI or a lexicon with is domain-adapted to the data set under consideration as a lexicon base for Lex2Sent.
\subsection{Lex2Sent as an initial fit}
While Lex2Sent is designed for a low hardware resource environment without a GPU, it can still be benefitial to use it in combination with larger, pre-trained models like RoBERTa. To demonstrate this, we use Lex2Sent's beneficial property of displaying a degree of certainty in its results based on how high or low the value of $\textit{diff}_d^{\text{mean}}$ is for $d=1,\dotsc,D$. To create data set for our RoBERTa model to fine-tune on, we therefore only use 10\% of the data set: the 5\% documents that have the highest and 5\% that have the lowest values of our training data set. We fine-tune this version of RoBERTa in 30 epochs using LoRA \cite{LoRA} with $r=8$ and thus 1,838,082 trainable parameters.

To evaluate this approach, we use the iMDb data set, as it contains both a training data set for Lex2Sent to train and RoBERTa to fine-tune on and a test data set for out-of-sample observations that can be classified by RoBERTa. We repeated this procedure five times. On average, our fine-tuned model classified 85.47\% of all test documents correctly. While this does not match GPT's classification rate, it does yield the advantage of being cost-efficient. This indicates that Lex2Sent can make for a good initial fit for an active learning approach. Starting from this classification rate, a human-in-the-loop style annotation might take place to improve the classification further.
\section{Conclusion}
\label{Summary}
Text classification is commonly performed in a supervised manner using a hand-labeled data set. Unsupervised classification can help when there is no such annotated data set available. This paper proposes the Lex2Sent model, which steers an intermediate course between learning-based and deterministic approaches to create an unsupervised classification, which can be created in a low hardware resource environment without access to a GPU. A binary lexicon is used as a replacement for the missing information that is usually represented by the annotations. The performance of this method is increased by aggregating the results from resampled data sets, which can be seen as a bagging effect.

Lex2Sent yields higher classification rates than all six analyzed sentiment lexica on all three data sets under study, no matter the lexicon-base. Our findings indicate that this might be caused by classifying documents in a more balanced way compared to traditional lexicon methods. Despite being a learning-based approach, the Lex2Sent method shows higher classification rates than traditional lexica on smaller data sets.
\section*{Ethical Considerations}
While our model requires calculating multiple Doc2Vec models for a single analysis, we modified our model specifications and the number of executions to keep the computational budget manageable in the context of climate change \citep{Strubbel}. Hence, we perform 50 executions in all of our experiments to ensure that the results are not affected by outliers, but the computational budget remains within reasonable boundaries. Our choice of using the fixed grid with 36 parameter combinations is also caused by this goal. Using this grid, each model finished training in less than two hours.
\section*{Limitations}
While Lex2Sent improves the classification rate of lexica, it is not capable of reaching the classification rates of models like GPT, but should be seen as a much less resource intensive alternative for the specific task of binary text classification. 

Lex2Sent's architecture is independent of the type of binary classification task at hand, so it should work similarly well for other classification tasks given suitable lexica. This is however a theoretical assumption, as we have tested Lex2Sent's capabilities for sentiment analysis specifically.

Lex2Sent has been designed for a two-label-case. To use it in a ordinally scaled multi-label-case, we would need to create multiple thresholds that determines the predicted class, instead of just one. This yields new challenges, as we can not heuristically choose the threshold as $0$ like in a binary classification task.

While Lex2Sent's architecture does not depend on the language of the documents or the lexica, it should theoretically perform just as well in low resource languages without needing large training data sets like sophisticated language models. We have not tested this hypothesis though.

\section*{Acknowledgments}
This paper is part of a project of the Dortmund Center for data-based Media Analysis (\href{https://docma.tu-dortmund.de/}{DoCMA}) at TU Dortmund University. The work was supported by the Mercator Research Center Ruhr (MERCUR) with project number Pe-2019-0044. In addition, the authors gratefully acknowledge the computing time provided on the Linux HPC cluster at TU Dortmund University (LiDO3), partially funded in the course of the Large-Scale Equipment Initiative by the German Research Foundation (DFG) as project 271512359.

\bibliography{Masterarbeit}
\end{document}